# Obstacle crossing strategies for high-speed 4WD small-scale vehicle

D. N'chot[1], R. Lenain[2], J.-C. Fauroux[1,3], L. Al Bassit[2] and P. Vaslin[4]

[1] Université Clermont Auvergne, Clermont Auvergne INP, CNRS, Institut Pascal, F-63000 Clermont-Ferrand, France., e-mail: denis.n-chot@sigma-clermont.fr

[2] Université Clermont Auvergne, INRAE, UR TSCF, F-63000 Clermont–Ferrand, France., e-mails: roland.lenain@inrae.fr ; lama.al-bassit@inrae.fr

[3] MecaBotiX, At Home / La Cité, 55 avenue Louis Bréguet, F-31400 Toulouse, France., e-mail: jc.fauroux@mecabotix.com

[4] Université Clermont Auvergne, Clermont Auvergne INP, EMSE, CNRS, LIMOS, F-63000 Clermont-Ferrand, France., e-mail: philippe.vaslin@limos.fr

**Abstract**

Unmanned ground vehicle obstacle crossing generally relies on two strategies: (i) applying a wheel torque for climbing and (ii) modifying the vehicle shape by using a wheel-leg or wheel-paddle to lift the wheel on top of the obstacle. However, most of those strategies sacrifice speed in order to have a longer contact duration between the wheels and the obstacle. This paper investigates the behaviour of a 4WD high-speed vehicle while crossing a step obstacle using a design of experiment (DoE). A 3D multibody vehicle model is equipped with a novel 2-DoF suspension system, which horizontal damping coefficient is modify to dampen wheel motion in longitudinal and vertical directions in relation to the chassis, for a given speed and obstacle height. The DoE results allow to propose a novel high-speed obstacle crossing strategy based on three metrics: (i) the kinetic energy variation of the vehicle, (ii) the contact duration between the wheel and the obstacle, and (iii) the pitch rate at the start of the ballistic phase. Experimental function are proposed to be able modify these metric in real time.

# 1 Introduction

Obstacle crossing of wheeled off-road vehicles starts from the contact between the wheel and the obstacle. When obstacle slope is sufficient, the horizontal component of the ground reaction force reaches the same order of magnitude as the vertical component [1], which creates dynamic pitch, roll and yaw momentum, resulting in a high vehicle speed reduction and an increasing risk of vehicle tipping over. Obstacle crossing generally relies on two strategies: (i) to apply sufficient wheel torque to climb and (ii) to modify the shape of the vehicle by using wheel + leg or wheel + blade, for instance [2, 3, 4], with or without passive or active suspension mechanism, to lift the wheels on top of the obstacle. These strategies sacrifice the vehicle speed [2] to maintain contact between the wheels and the obstacle. Indeed, the wheel continues to climb over the obstacle when torque and friction conditions are met [3]. In literature, the wheel radius is the maximum height of an obstacle that can be crossed by a vehicle [4, 5]. Previous work [1, 6] have shown that the Susp-4D suspension system with 2-DoF: (i) reduces the deceleration of a vehicle with a certain speed over an obstacle (ii) overcomes obstacles of greater height than the wheel radius and (iii) reduces the risk of tipping over. This paper proposes a study of a step obstacle crossing at high speed by a 4WD vehicle equipped with Susp-4D to discuss novel strategies based on the control of the horizontal suspensions to optimize obstacle crossing trough three criteria:

- to save kinetic energy of the vehicle;
- to have a contact duration long enough to climb the obstacle;
- to reduce the pitch rate at the start of the ballistic phase.

Section 2 presents the multi-body model of the vehicle and the design of experiment. Sections 3, 4 and 5 present the results of the experimental design, discuss the control strategy and finally conclude with some future work.

## 2 Material and methods

### 2.1 Multi-body model description of the full vehicle

The Adams multi-body modelling software was used to make a 3D model of a 4WD vehicle (Fig. 1). The vehicle consists of a rectangular chassis and four driven wheels. The chassis is constrained with a plane linkage (P2: Fig. 2(b)) to have a 2D crossing. Both front wheel are connected to the chassis with Susp-4D, whereas both rear wheels are connected to the chassis with Susp-3D suspensions, which are Susp-4D with locked horizontal suspensions. The two first possible motions of front wheels are longitudinal and vertical with respect to the chassis, while rear wheels can only move vertically, according to the kinematics (Fig. 2(a)) of Susp-4D and Susp-3D [6], respectively. The other two degrees of freedom of the wheel are rotations around Yo and around Zo (Fig. 1(a)). The obstacle is a rectangular bar of height (hO) fixed to the ground. The vehicle centre of gravity ($Z_{CoG}$) is at 0.13 m and the wheel radius (wr) is at 0.0745 m.

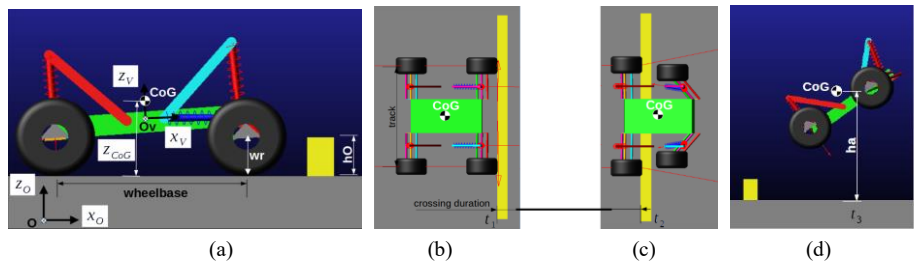

(a)　　　　　(b)　　　　　(c)　　　　　(d)

**Figure 1** 3D model of the vehicle (a) before, (b) during and (c) after the obstacle crossing

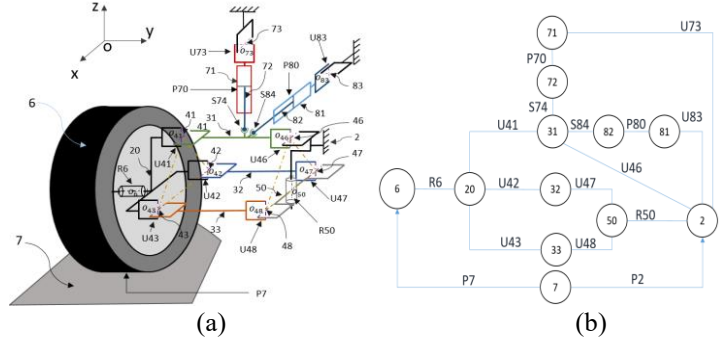

(a) (b)

**Figure 2** (a) Kinematic diagram and (b) linkage graph of Susp4D. The markers and nomenclature of the solids are given as follows: 2 : chassis ; 6 : wheel ; 7 : ground ; 20 : hub carrier ; 31 : upper shaft; 32 : rear direction shaft; 33 : front direction shaft; 41, 42, 43, 46, 47, 48, 73, 83 : cross-bar; 50 : rudder; 71 : vertical body of cylinder; 72 : vertical rod of cylinder ; 81 : longitudinal body of cylinder;  82 : piston of cylinder; And here the joints: Po7 : ponctual linkage; P70, P80 : prismatic linkages; R6, R50 : pivot linkages; S74, S84 : spherical linkages; U41, U42, U43, U46, U47, U48, U73, U83 : universal joint.

The vehicle tyre is made with an outer rubber envelope filled with elastic foam and the ground is asphalt. The wheel-obstacle contact model is based on the Hertzian contact theory, which use a spring damper formalism (Fig. 3(a)), and corresponds to IMPACT function of Adams:

$$F_{IMPACT(t)} = \begin{cases} 0 \text{ if } wr_1 > wr \\ k(wr - wr_1)^e - c_{max}\dot{wr_1} * STEP(wr_1, wr - d, 1, wr, 0) \text{ if } wr_1 \leq wr \end{cases} \quad (1)$$

The stiffness value (k) of the contact force takes into account both the material properties and the contact geometry, and ensures the conformity of the simulation with the physical phenomenon [7]. The literature proposes several stiffness values for rubber [8, 9]. the chosen value allows the most realistic simulation with respect of the physics. It is obtaned by a trial and error approach.

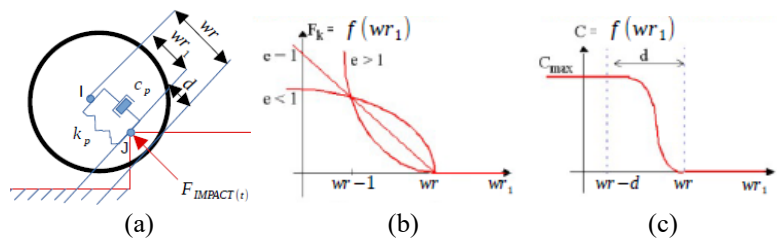

(a) (b) (c)

**Figure 3** Wheel-ground contact model in Adams (a) impact force (b) exponent selection charts (c) evolution of the damping coefficient.

The exponent (e) of the stiffness force depends on the material. The value given in the Table 1 corresponds to a good value for rubber. The recommended damping value (cmax) [7] in Adams is around 1% of the stiffness value. It is chosen to guarantee the penetration depth. The static and dynamic coefficients are chosen among existing values in the literature [10] and tuned to allow realistic physical behaviour in simulation. The stiction values are obtained using the same method as friction ones.

**Table 1.** Wheel-ground contact parameters used in Adams simulations.

| Stiffness (k) N/mm | Force Exponent (e) | Damping (cmax) Ns/mm | Penetration deph (d) | Static coefficient ($\mu s$) | Dynamic coefficient ($\mu d$) | Stiction Transition Vel. | Friction Transition Vel. |
|---|---|---|---|---|---|---|---|
| 1000 | 1.1 | 10 | 0.01 | 1 | 0.95 | 1500 | 4000 |

## *2.1 Design Of Experiment for obstacle crossing*

The obstacle crossing can be divided into three phases: (1) Before the crossing: all four wheels of the vehicle are on the ground and rolling without slipping, the vehicle is moving towards the obstacle (Fig. 1(a)); (2) During the crossing: the front wheels start from the pre-constrained position ($C_p$), the crossing starts at time ($t_1$) when the front wheels collide with the obstacle (Fig. 1(b)), the collision phase ends when the wheels reach the maximum longitudinal displacement, the crossing ends at time ($t_2$) after the rear wheel has collided with the obstacle or passed over the obstacle without touching it (Fig. 1(c)); (3) After the crossing: the vehicle reaches the peak of its trajectory at time ($t_3$) before falling back to the ground (Fig. 1(d)).

This paper aims at studying the influence of the influence of the suspension parameters of the proposed Susp-4D suspension to optimize obstacle crossing. As consequence, we consider three metrics that qualifies the crossing: - the kinetic energy variation ($\Delta E_c$) of its maximum and minimum value between ($t_1$) and ($t_2$) to reduce the speed loss; - the pitch rate ($\dot{\delta}$) at the end of the crossing ($t_2$) to reduce the pitch rate and finally; the wheel-obstacle contact duration (CDWO) to climb over the obstacle. The contact duration is the time taken by the wheel to reach the maximum longitudinal shortening.

The parameters chosen for the design of experiment (Fig. 4) are from three categories: geometric parameter: obstacle height (hO); kinematic parameter: initial vehicle speed (vc) and kinetostatic parameter: front longitudinal damping coefficient ($c_{AV}$).

We choose 5 obstacle heights (hO), 5 vehicle speeds (vc) and finally 5 damping coefficients of the front longitudinal suspensions ($c_{AV}$), which corresponds to 243 trials (Table 1). The torque applied to the wheels ($\tau$) is monitored to maintain a constant speed throughout the tests before crossing and the stiffness (k) of the suspension is set to 2.26 N/mm.

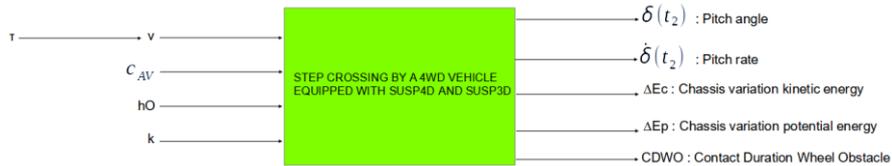

**Figure 4** Synoptic of the experiment design.

**Table 1.** Numerical value of the different parameters, with wr = 0.0745 m.

| Parameters | | | | | |
|---|---|---|---|---|---|
| hO (m) | 25%wr | 50%wr | 80%wr | 90%wr | 100%wr |
| vc (m/s) | 3 | 6 | 9 | 12 | 15 |
| $c_{AV}$ (Ns/mm) | 0.4 | 0.8 | 1.6 | 3.2 | 6.4 |

# 3 Results of the experiment design

## *3.1 Influences of the damper coefficient when crossing*

Let us first consider an obstacle crossing of 25% of the wheel radius at a constant speed of 12 m/s for different damping coefficients (0.4, 1.6 and 6.4). Three simulations are then run without modifying other parameters. On each curve, the successive points correspond to instants t1, t2 and t3, in chronological order of crossing. During the crossing phase between (t1) and (t2), the relative velocity and kinetic energy loss (Fig. 5(a) and (b) resp.); the pitch angle and pitch rate at time $t_2$ (Fig. 6(a) and (b)); the chassis acceleration at time $t_1$ (Fig. 7(a)) decrease, the more the damping coefficient is low. For the maximum torque applied (Fig. 7(b)); the vehicle height is at apogee (Fig. 8(a)) and the horizontal displacement of the wheel in relation to the chassis (Fig. 8(b)) increase, the more the damping coefficient is low.

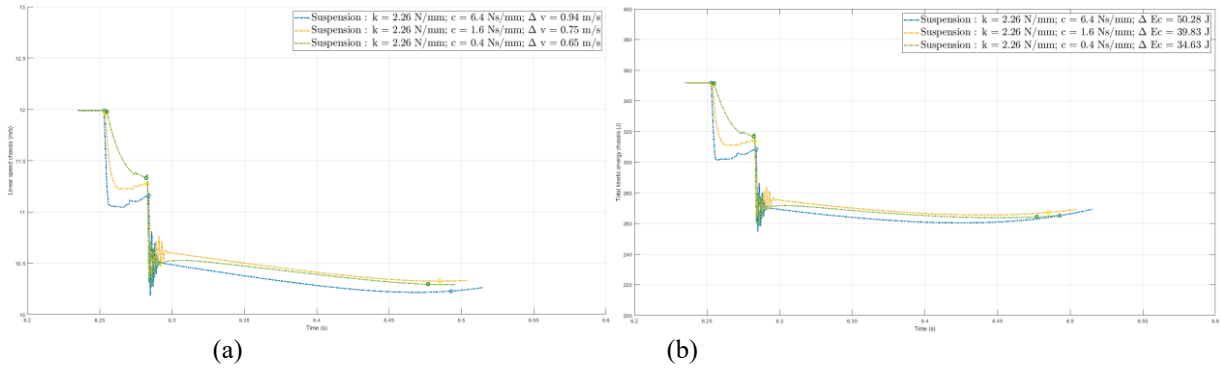

(a)                                           (b)

**Figure 5** Evolution of (a) the chassis speed vc and (b) the chassis total kinetic energy (Ec).

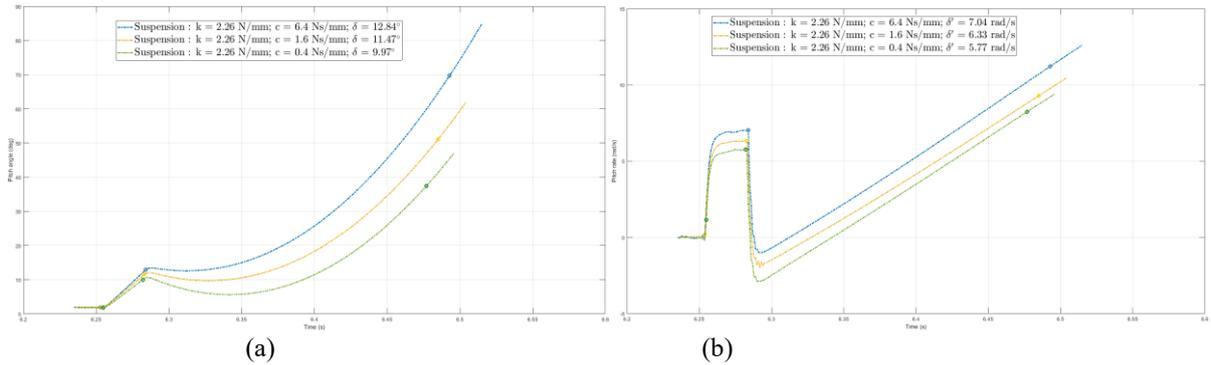

(a)                                           (b)

**Figure 6** Evolution of (a) the chassis pitch angle and (b) the chassis rate during the crossing.

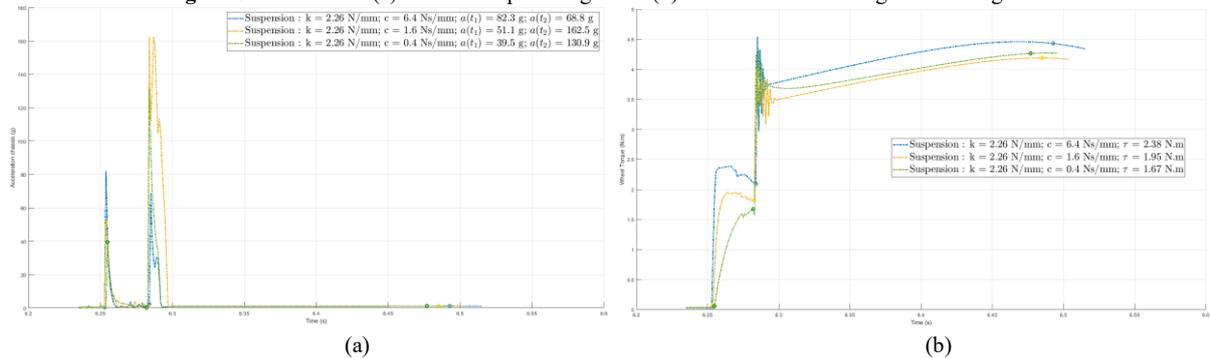

(a)                                           (b)

**Figure 7** Evolution of (a) chassis acceleration and (b) torque applied to the wheels during the crossing.

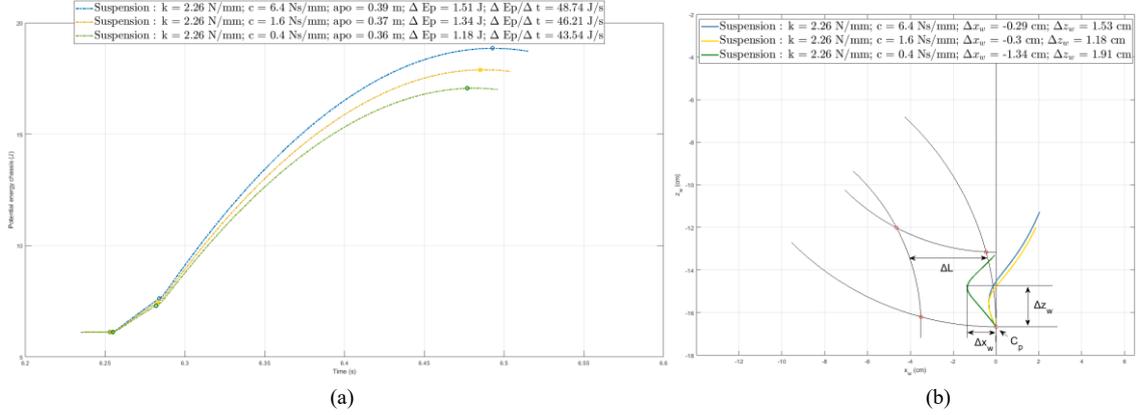

(a)                                                                  (b)

**Figure 8** Evolution of (a) the potential energy of the chassis and (b) the trajectory of the wheel on the workspace of the mechanism during obstacle crossing.

## 3.2 Influence of parameters on each metrics when crossing

In this section, we study the influence of obstacle height and vehicle speed for several values of damping coefficient. Several simulations are achieved with the parameters values of Table 1, which leads to 243 trials, in order to determine a mathematical relationship linking the chosen metrics to parameters. Such relationships are depicted on figures 8(a), (b) and (c) for the three metrics chosen ($\Delta E_c$, $\dot{\delta}$, CDWO), respectively, at different obstacle height which correspond to cloud point with different colors (Fig. 8).

The obstacle height (hO), the speed (vc), the damping coefficient ($c_{AV}$), influences the kinetic energy variation ($\Delta E_c$) in the same direction. The higher hO is, for a given vc and $c_{AV}$; the higher vc is, for a given hO and $c_{AV}$; or the higher the $c_{AV}$ is, for a given vc and hO, the higher ($\Delta E_c$) is (Fig. 8(a)).

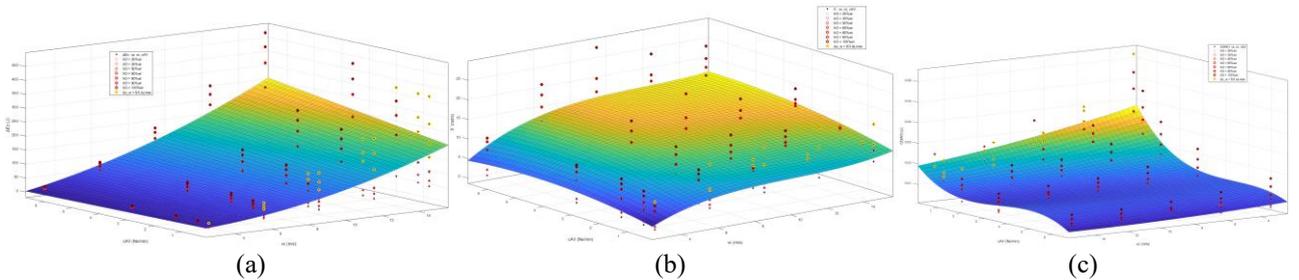

(a)                              (b)                              (c)

**Figure 9** Fitting function of the responses chosen (a) Kinetic energy variation (b) Pitch rate (c) Wheel Obstacle Contact Duration. With yellow dot for $\Delta x\_w > 8/3\ \Delta L_{max}$ and $\Delta L$ correspond to the mechanism longitudinal displacement.

The obstacle height (hO), the speed (vc) influences the pitch rate value ($\dot{\delta}$) in the same direction. The higher (hO) is for a given vc and $c_{AV}$; the higher vc is for a given hO and $c_{AV}$, the higher the pitch rate is at time $t_2$ and vice versa. The damping coefficient ($c_{AV}$) influences ($\dot{\delta}$) in a parabolic way for a given (vc) and (hO). The maximum pitch rate value lies in the range of the damping coefficient between 3.2 and 6.4 (Fig. 8(b)).

The obstacle height (hO) influences the obstacle wheel contact duration (CDWO) in the same direction. The higher (hO) is for a given vc and $c_{AV}$, the higher (CDWO) is (Fig. 8(c)). The speed (vc) influences (CDWO), in the opposite direction, at a given (hO), for ($c_{AV}$) values higher than 3.2 Ns/mm, on the one hand. In this case, the lower (vc), the higher CDWO is. On the other hand, for ($c_{AV}$) lower than 3.2 Ns/mm, in a parabolic way. In this case, (CDWO) is higher for extreme speeds (3 and 15 m/s) and reaches its minimum value for intermediate speeds. The damping coefficient ($c_{AV}$) influences (CDWO) in the opposite direction for a given (vc) and (hO), (CDWO) increases as the ($c_{AV}$) decreases. We can propose experimental function that link the metrics to (vc) and ($c_{AV}$) for a given (hO):

$$\Delta E_c(v_c, c_{AV}) = e_{00} + e_{10}v_c + e_{01}c_{AV} + e_{20}v_c^2 + e_{11}v_c c_{AV} \quad (2)$$

With the values of $e_{00}$, $e_{10}$, $e_{01}$, $e_{20}$, $e_{11}$ changing according to the obstacle height.

$$\delta(\tau, c_{AV}) = p_{00} + p_{10}v_c + p_{01}c_{AV} + p_{20}v_c^2 + p_{11}v_c c_{AV} + p_{02}c_{AV}^2 + p_{30}v_c^3 + p_{21}v_c^2 c_{AV} + p_{12}v_c c_{AV}^2 \quad (3)$$

With the values $p_{00}$, $p_{10}$, $p_{01}$, $p_{20}$, $p_{11}$, $p_{02}$, $p_{30}$, $p_{21}$, $p_{12}$ changing according to the obstacle height.

$$CDWO(v_c, c_{AV}) = d_{00} + d_{10}v_c + d_{01}c_{AV} + d_{20}v_c^2 + d_{11}v_c c_{AV} + d_{02}c_{AV}^2 + d_{21}v_c^2 c_{AV} + d_{12}v_c c_{AV}^2 + d_{03}c_{AV}^3 \quad (4)$$

With the values $d_{00}$, $d_{10}$, $d_{01}$, $d_{20}$, $d_{11}$, $d_{02}$, $d_{21}$, $d_{12}$, $d_{03}$ changing according to the obstacle height.

# 4 2D strategies for reducing kinetic energy loss and improving vehicle stability during obstacle crossing

The anticipatory strategy of obstacle crossing is based on the detection of the obstacle height to adjust the damping coefficient and torque parameters. Before the vehicle is crossing, the damping coefficient is set to its maximum value to limit the load transfer of the chassis during acceleration. At a maximum speed (vc) of 15 m/s and minimum obstacle detection distance of 1 m. There is only 0.067 s left to apply sufficient torque ($\tau$) and reduce the damping coefficient ($c_{AV}$).

At this speed, during the crossing, the vehicle takes off the ground regardless of the applied torque and damping coefficient. In this case, it is necessary to apply an adapted torque at $t_1$ and $t_2$, for reducing the pitch rate of the chassis.

The novel approach of high-speed obstacle crossing rely on an optimisation problem whose objectives is to minimise ($\Delta E_c$) during crossing while maximising (CDWO) in order to apply sufficient torque during the crossing that minimises the pitch rate at the end of the crossing ($t_2$). The problem can be express mathematically as follow:

To minimise $\Delta E_c(v_c, c_{AV})$, $\delta(\tau, c_{AV})$ ; to maximise $CDWO(v_c, c_{AV})$ ;

With the following contraints :
- The damping coefficient must be chosen in such a way as to avoid reaching the stops of the suspensions and to keep the wheel centre horizontal displacement in line (Fig. 9).

$$\Delta x_w < \frac{8}{3}\Delta L \quad (4)$$

- The torque should be sufficient to climb, but not so strong as to cause the tire to rip on the obstacle:

$$\tau_{min} \leq \tau \leq \tau_{max} \quad (5)$$

In addition, there are constraints related to the equipment that will reduce the parameters research area. The stroke of the cylinders limit the horizontal displacement of the wheel centre. The motor of the vehicle limits the maximum torque. The dimensions of the cylinder and characteristics of the valve limit in minimum and maximum value the damping coefficient.

## 5 Conclusion

We studied in simulation the behaviour of a 3D multi-body 4WD vehicle during a step obstacle crossing with different obstacle height (hO) at different speeds (vc) and with different damping coefficients ($c_{AV}$) of the horizontal suspensions. These three parameters with their five levels chosen for the Design of Experiment (DoE) leads to 243 trials. We qualified the crossing by three metrics: the kinetic energy variation ($\Delta E_c$), the pitch rate ($\dot{\theta}$) at ballistic phase start $t_2$ and the wheel obstacle contact duration (CDWO).

In all trials, a lower damping coefficient results in less loss of speed, kinetic energy variation, lower pitch rate at $t_2$, and a longer wheel obstacle contact duration. This DoE enable to find an empiric relationship between parameters and metrics, which will help, in future work, to develop a robust control law of the longitudinal suspensions for better obstacle crossing.

**Acknowledgments** This work has been sponsored by the French government research program "Investissements d'Avenir" through the IDEX-ISITE initiative 16-IDEX-0001 (CAP 20-25) and the IMobS3 Laboratory of Excellence (ANR-10-LABX-16-01).

## References

1. Fauroux, J.C., Bouzgarrou, B.C.: dynamic obstacle-crossing of a wheeled rover with double wishbone suspension. In: Field Robotics. pp. 642-649. (2011). Last seen 30/08/22
http://jc.fauroux.free.fr/PUB/ARTICLES/2011_CLAWAR_Fauroux_Bouzgarrou_DRAFT_Dynamic_obstacle-crossing_of_a_wheeled_rover_with_double_wishbone_suspension.pdf
2. Lauria, M. Nouveaux concepts de locomotion pour véhicules tout-terrain robotisés. Thesis EPFL. (2003)
3. Abad-Manterola, P., Burdick, J.W., Nesnas, I.A.D., Chinchali, S., Fuller, C., Xuecheng Zhou: Axel rover paddle wheel design, efficiency, and sinkage on deformable terrain. In: 2010 IEEE International Conference on Robotics and Automation. pp. 2821-2827. (2010)
4. Shivesh Kumar: All-terrain mobile robot for extra-terrestrial applications. Bachelor dissertation, 50 p. (2012). Last seen 30/08/22
https://doi.org/10.13140/RG.2.1.2331.1449
5. Wada, M.: Studies on 4WD Mobile Robots Climbing Up a Step. In: 2006 IEEE International Conference on Robotics and Biomimetics. p. 1529-1534. (2006)
6. N'chot, D., Fauroux, J.-C., Al Bassit, L., Lenain, R., Vaslin, P.: Performance Evaluation of an Innovative Suspension System for Obstacle Crossing. In: Cascalho, J.M., Tokhi, M.O., Silva, M.F., Mendes, A., Goher, K., and Funk, M. (éd.) Robotics in Natural Settings. pp. 95-106. (2023)
7. Verheul, C., International, S.: Benelux ADAMS User Meeting.19 p. (2012)
8. Gregory, M.J.: Measurement of rubber properties for design. Polymer Testing. Vol 4, pp. 211-223 (1984). Last seen 30/08/22 https://doi.org/10.1016/0142-9418(84)90013-8
9. Yu, W., Zhijun, L., Baolin, L., Mingshuai, G.: 1180. Stiffness characteristic comparison between metal-rubber and rubber isolator under sonic vibration. Vol 16, Issue 11 (2014)
10. J.Y. WONG. Theory of ground vehicles. John Wiley & Sons. 4th Edition. 2008.
https://www.engineersedge.com/coeffients_of_friction.htm